\documentclass{article} 

\usepackage{nfam2025_workshop}  
\usepackage{times}              

\usepackage{amsmath,amsfonts,bm}

\def\eqref#1{equation~\ref{#1}}

\def\1{\bm{1}}

\DeclareMathAlphabet{\mathsfit}{\encodingdefault}{\sfdefault}{m}{sl}
\SetMathAlphabet{\mathsfit}{bold}{\encodingdefault}{\sfdefault}{bx}{n}

\usepackage[backref=page]{hyperref}
\usepackage{url}
\usepackage{graphicx}
\usepackage{subcaption}

\title{Effects of Feature Correlations on Associative Memory Capacity}

\author{Stefan Bielmeier\thanks{Research conducted while a Visiting Research Scholar at UC Berkeley.}\\
  Department of Computer Science and Electrical Engineering\\
  University of California, Berkeley\\
  Berkeley, CA 94720 \\
  \texttt{stefan.bielmeier@tum.de} \\
\AND
  Gerald Friedland\thanks{Work done prior to joining Amazon.} \\
  Department of Computer Science and Electrical Engineering\\
  University of California, Berkeley \\
  Berkeley, CA 94720 \\
  \texttt{fractor@icsi.berkeley.edu} \\
}

\iclrfinalcopy

\begin{document}

\maketitle

\begin{abstract}
We investigate how feature correlations influence the capacity of Dense Associative Memory (DAM), a Transformer attention-like model. Practical machine learning scenarios involve feature-correlated data and learn representations in the input space, but current capacity analyses do not account for this. We develop an empirical framework to analyze the effects of data structure on capacity dynamics. Specifically, we systematically construct datasets that vary in feature correlation and pattern separation using Hamming distance from information theory, and compute the model's corresponding storage capacity using a simple binary search algorithm. Our experiments confirm that memory capacity scales exponentially with increasing separation in the input space. Feature correlations do not alter this relationship fundamentally, but reduce capacity slightly at constant separation. This effect is amplified at higher polynomial degrees in the energy function, suggesting that Associative Memory is more limited in depicting higher-order interactions between features than patterns. Our findings bridge theoretical work and practical settings for DAM, and might inspire more data-centric methods.
\end{abstract}

\section{Introduction}
Hopfield networks have gained renewed attention because they exhibit parallels with Transformer architectures and demonstrate strong empirical performance in diverse applications \citep{vaswaniAttentionAllYou2017, ramsauerHopfieldNetworksAll2020, widrichModernHopfieldNetworks2020}. Originally, their fundamental ability to store and retrieve patterns was limited to a fraction of the number of its neurons \citep{hopfieldNeuralNetworksPhysical1982a, hopfieldNeuronsGradedResponse1984}. Researchers recently expanded its storage capabilities up to exponentially many memories as Dense Associative Memory (DAM) \citep{dams_krotov_hopfield, demircigilModelAssociativeMemory2017}, and simplified their integration into modern machine learning architectures for practical problem solving \citep{ramsauerHopfieldNetworksAll2020}.

However, capacity dynamics under real-world data remain under-explored. While we know that the distribution of memories, and pattern separation in particular, affects storage limits \citep{wuUniformMemoryRetrieval2024}, most capacity analyses assume theoretically convenient distributions of nearly ideally-separated patterns with independent variables \citep{dams_krotov_hopfield, demircigilModelAssociativeMemory2017, ramsauerHopfieldNetworksAll2020}. This assumption does not hold for practical settings, and real-world data capacity already deviates in part from theoretical predictions \citep{chaudhryLongSequenceHopfield2023}. Therefore understanding how Hopfield models behave in data scenarios with imperfect separation and codependent features is an open question.

To address this, we examine Dense Associative Memory (DAM) in raw input space---mirroring scenarios where feature dimensions are fixed---to reflect practical machine learning contexts \citep{kempeNewFrontiersAssociative2024}. We develop a rigorous framework to systematically measure DAM capacity under various data scenarios. Our approach generates patterns with controlled separations to isolate the role of feature correlations in memory storage and retrieval, and measures memory capacity for various model configurations. This work offers insights into capacity behavior that complement existing theoretical analyses.

\section{Related Work and Background}
\textbf{Dense Associative Memory.}
Hopfield networks are a model of auto-associative memory, which is able to recover learned patterns from disrupted versions of themselves \citep{hopfieldNeuralNetworksPhysical1982a}. Dense Associative Memory (DAM) is a variant of it that has a rectified polynomial of degree $n$ as its activation function, and large storage capacity that increases with $n$ \citep{dams_krotov_hopfield}.

\textbf{Memory Capacity.} This is the number of patterns the network can store and retrieve reliably at a given error probability \citep{friedlandInformationDrivenMachineLearning2024}. Our target is $P_{error} = 0$. We follow common practice and initialize the network with each pattern in question and perform one asynchronous update to check retrieval performance \citep{lucibelloExponentialCapacityDense2024, ramsauerHopfieldNetworksAll2020, demircigilModelAssociativeMemory2017}.

\textbf{Kernel Capacity Methods.} Recent work exploits kernel transformations to increase Hopfield-like model capacity by making patterns more separable in a high-dimensional feature space via either the energy function of the model, or by reshaping the dataset directly \citep{wuUniformMemoryRetrieval2024, huProvablyOptimalMemory2024}. These techniques presuppose freedom to alter data and semantic representations---an assumption that does not always hold in practical settings with fixed input dimensions, such as raw images.

\textbf{Pattern Separation and Hamming distance.} 
Higher average pairwise pattern separation increases memory capacity, and Hamming distance can be used to identify and remove overlapping patterns to do the same \citep{huProvablyOptimalMemory2024, huSparseModernHopfield2023, manandharEffectHammingDistance2002}. We view stored patterns as Hamming codes from information theory \citet{hammingErrorDetectingError1950, shannonMathematicalTheoryCommunication1948}. For a formal definition of Hamming distance (HD) in context: given two bitstrings ${\zeta_1}$ and $\zeta_2$, the Hamming distance between them counts the number of differing bit positions. In our framework, this metric is especially useful because each bit difference translates into a discrete, uniform step of dissimilarity, making it easier to design construction algorithms that control average separation.

\section{Method and Experimental Setup}
We want to measure memory capacity for a set of patterns of different separation. For this, we need to (1) create datasets for both synthetic and real data for a given level of average separation to isolate effects of feature correlations, and (2) compute memory capacity efficiently.

\textbf{Dataset Construction.} We create two main datasets to compare capacity dynamics. The first dataset is the baseline, an "artificial" dataset that contains 50 subsets of skewed i.i.d. Rademacher patterns. This distribution is parameterized by a single parameter which determines whether we sample $1$ or $-1$. We vary the parameter between (0.5, 1] in steps of $eps = 0.01$, resulting in a dataset of size $50 \times \ 50 \times \ 784$. The average Hamming distance consequently ranges from $(0,392)$. The second dataset consists of binarized MNIST images at threshold 128. We use a greedy selection procedure---an iterative hill-climbing algorithm---to ensure a specified Hamming distance in each subset. Specifically, we randomly sample a pattern and accept it into our subset only if it increases or maintains the target HD across all previously accepted patterns. Since data structure inherent to MNIST limits average separation in subsets to $30 \leq HD \leq 190$, we create 53 subsets to evenly space out data points. This results in a dataset of size $53 \times 50 \times 784$.

\textbf{Model Setup.} We fix the number of neurons at  $N=784$ for a $28 \times 28$ input size, and use a rectified polynomial as in \citet{dams_krotov_hopfield}, with $n$ controlling the polynomial degree. We compute capacities for $n \in [6, 38]$ with step size $2$ for values larger than 11. Our program also stores input patterns in-memory instead of training weights for simplicity and speed.

\textbf{Capacity Measurement.} We would need to train and test each model configuration on all possible datasets, which is very resource-intensive. In practice, we can only test a representative subset that we need to identify: given a constructed subset of size $S$, we apply binary search to find memory capacity $K_{max}$, which is the largest $K < S$ such that our DAM can store and retrieve the first $K$ patterns perfectly (normalized inner product = 1). This method reduces computational costs from $O(s)$ to $O(log \ s)$, $s$ being the number of evaluated $K$s. Importantly, we re-calculate the mean separation after we determine $K_{max}$ to accurately represent separation in plots. We also exclude values of $K_{max} > 49$ from results to avoid distortions: since all subsets contain 50 patterns $S = 50$, memory capacity $K_{max}$ saturates at $50$ for high polynomial degrees of $n$ even though the actual $K_{max}$ might be larger than 50. Lastly, we also use early stopping based on $K_{max}$ thresholds to optimize computational efficiency.

\section{Results}
\begin{figure}[ht]
    \vspace{-1.5\baselineskip} 
    \centering   
    \includegraphics[width=\linewidth]{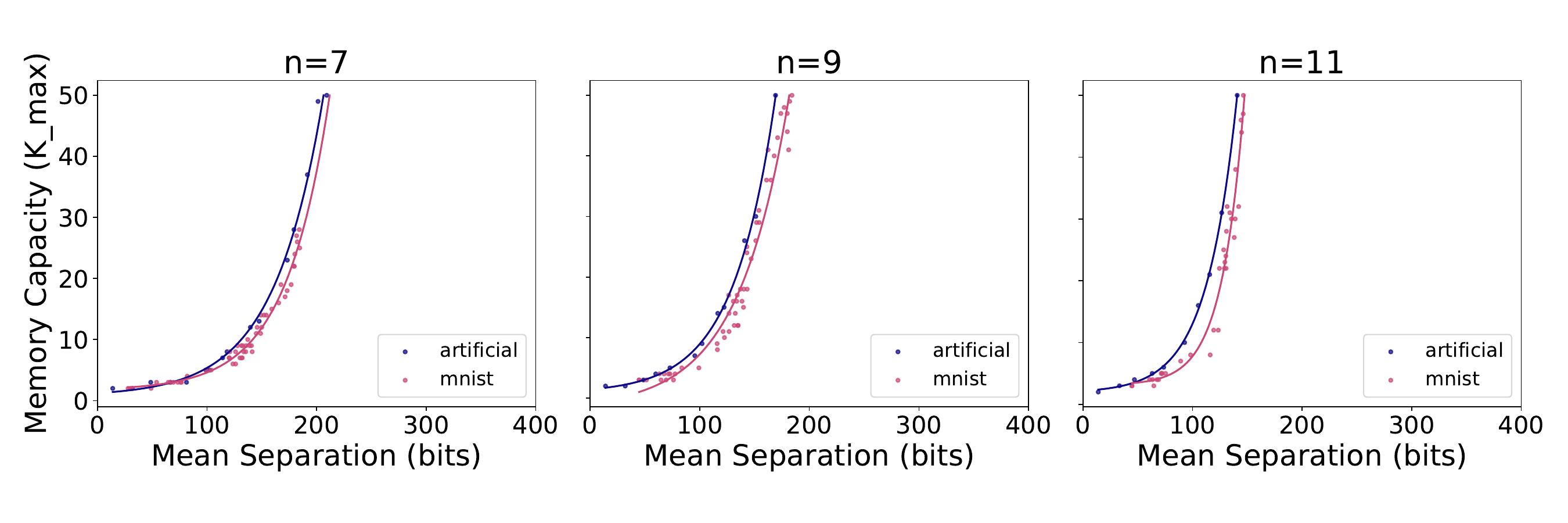}
    \captionsetup{skip=-4pt}
    \caption{Comparing memory capacity scaling for synthetic and real data. We observe a slightly increasing constant difference between them at various levels of $n$.}
    \label{fig:comp}

    \includegraphics[width=\linewidth]{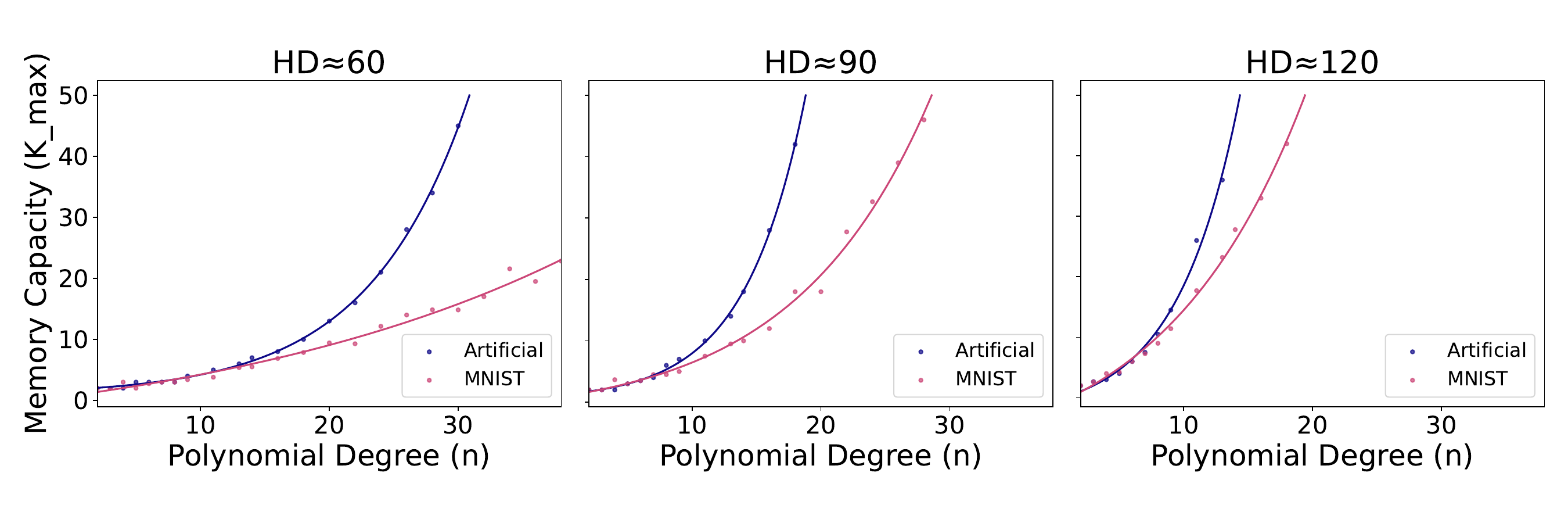}
    \captionsetup{skip=-4pt}
    \caption{Capacity scaling for increasing $n$ at different Hamming distance buckets.}
    \label{fig:buckets}
\end{figure}

As a baseline analysis, we confirm empirically that memory capacity $K_{\max}$ scales exponentially with the average separation in the dataset, shown in Appendix \ref{appendix:baseline}. In Figure \ref{fig:comp}, we fix $n$ and sweep the mean separation. As expected, capacity $K_{\max}$ grows sharply with separation, but synthetic data achieves consistently higher $K_{\max}$ for a given separation than MNIST. This effect becomes slightly more apparent at higher $n$. To better understand this dynamic, we plot $K_{max}$ against polynomial degree $n$ for both synthetic (blue) and MNIST (red) data across different mean Hamming distance buckets in Figure \ref{fig:buckets}. Buckets have a HD leeway of 10 on both sides. 

As expected, capacity grows exponentially with $n$. While pattern separation primarily drives capacity, correlated data is disadvantaged at higher $n$. At higher polynomial degrees, correlated data (MNIST) notably suffers an increasing capacity drop relative to synthetic data. Higher separation somewhat mitigates the strength of this divergence, but its effect is substantial for the the practical dataset in question: the MNIST train set with 3000 of each digit has an avg. $HD \approx 112$, where divergence for large $n$ is still significant. 

\section{Discussion and future directions}
With this paper, we bridge the gap between theoretical analyses and nuanced realities of real data. Our results highlight that feature correlations affect memory capacity of Hopfield networks significantly and are relevant for practical settings. Ignoring feature correlations can overestimate real-world capacity. These observations suggest that Transformer architectures, and by extension Large Language Models (LLMs), could face similar constraints when processing correlated inputs, a hypothesis worth exploring in future work.

We leave a rigorous theoretical explanation and intuition of how these effects come about to future work. Another, more immediate next research step could be how continuous-space input data affects capacity dynamics. It would also be interesting to study bounds below optimal memorization and more toward generalization, or recall under the assumption of non-zero error rates, both of which usually happen in practical machine learning scenarios.

\bibliography{main}

\begin{thebibliography}{17}
\providecommand{\natexlab}[1]{#1}
\providecommand{\url}[1]{\texttt{#1}}
\expandafter\ifx\csname urlstyle\endcsname\relax
  \providecommand{\doi}[1]{doi: #1}\else
  \providecommand{\doi}{doi: \begingroup \urlstyle{rm}\Url}\fi

\bibitem[Chaudhry et~al.(2023)Chaudhry, {Zavatone-Veth}, Krotov, and Pehlevan]{chaudhryLongSequenceHopfield2023}
Hamza~Tahir Chaudhry, Jacob~A. {Zavatone-Veth}, Dmitry Krotov, and Cengiz Pehlevan.
\newblock Long {{Sequence Hopfield Memory}}, November 2023.

\bibitem[Demircigil et~al.(2017)Demircigil, Heusel, L{\"o}we, Upgang, and Vermet]{demircigilModelAssociativeMemory2017}
Mete Demircigil, Judith Heusel, Matthias L{\"o}we, Sven Upgang, and Franck Vermet.
\newblock On a {{Model}} of {{Associative Memory}} with {{Huge Storage Capacity}}.
\newblock \emph{Journal of Statistical Physics}, 168\penalty0 (2):\penalty0 288--299, July 2017.
\newblock ISSN 1572-9613.
\newblock \doi{10.1007/s10955-017-1806-y}.

\bibitem[Friedland(2024)]{friedlandInformationDrivenMachineLearning2024}
Gerald Friedland.
\newblock \emph{Information-{{Driven Machine Learning}}: {{Data Science}} as an {{Engineering Discipline}}}.
\newblock Springer International Publishing, Cham, 2024.
\newblock ISBN 978-3-031-39476-8 978-3-031-39477-5.
\newblock \doi{10.1007/978-3-031-39477-5}.

\bibitem[Hamming(1950)]{hammingErrorDetectingError1950}
Richard~W. Hamming.
\newblock Error detecting and error correcting codes.
\newblock \emph{The Bell system technical journal}, 29\penalty0 (2):\penalty0 147--160, 1950.

\bibitem[Hopfield(1982)]{hopfieldNeuralNetworksPhysical1982a}
J~J Hopfield.
\newblock Neural networks and physical systems with emergent collective computational abilities.
\newblock \emph{Proceedings of the National Academy of Sciences}, 79\penalty0 (8):\penalty0 2554--2558, April 1982.
\newblock \doi{10.1073/pnas.79.8.2554}.

\bibitem[Hopfield(1984)]{hopfieldNeuronsGradedResponse1984}
J~J Hopfield.
\newblock Neurons with graded response have collective computational properties like those of two-state neurons.
\newblock \emph{Proceedings of the National Academy of Sciences}, 81\penalty0 (10):\penalty0 3088--3092, May 1984.
\newblock ISSN 0027-8424, 1091-6490.
\newblock \doi{10.1073/pnas.81.10.3088}.

\bibitem[Hu et~al.(2023)Hu, Yang, Wu, Xu, Chen, and Liu]{huSparseModernHopfield2023}
Jerry Yao-Chieh Hu, Donglin Yang, Dennis Wu, Chenwei Xu, Bo-Yu Chen, and Han Liu.
\newblock On sparse modern hopfield model.
\newblock \emph{Advances in Neural Information Processing Systems}, 36:\penalty0 27594--27608, 2023.

\bibitem[Hu et~al.(2024)Hu, Wu, and Liu]{huProvablyOptimalMemory2024}
Jerry Yao-Chieh Hu, Dennis Wu, and Han Liu.
\newblock Provably {{Optimal Memory Capacity}} for {{Modern Hopfield Models}}: {{Transformer-Compatible Dense Associative Memories}} as {{Spherical Codes}}, October 2024.

\bibitem[Kempe et~al.(2024)Kempe, Krotov, Kuehne, Lee, and Solla]{kempeNewFrontiersAssociative2024}
Julia Kempe, Dmitry Krotov, Hilde Kuehne, Daniel Lee, and Sara~A. Solla.
\newblock New {{Frontiers}} in {{Associative Memories}}.
\newblock In \emph{{{ICLR}} 2025 {{Workshop Proposals}}}, December 2024.

\bibitem[Krotov \& Hopfield(2016)Krotov and Hopfield]{dams_krotov_hopfield}
Dmitry Krotov and John~J Hopfield.
\newblock Dense associative memory for pattern recognition.
\newblock \emph{Advances in neural information processing systems}, 29:\penalty0 1172--1180, 2016.

\bibitem[Lucibello \& M{\'e}zard(2024)Lucibello and M{\'e}zard]{lucibelloExponentialCapacityDense2024}
Carlo Lucibello and Marc M{\'e}zard.
\newblock Exponential {{Capacity}} of {{Dense Associative Memories}}.
\newblock \emph{Physical Review Letters}, 132\penalty0 (7):\penalty0 077301, February 2024.
\newblock \doi{10.1103/PhysRevLett.132.077301}.

\bibitem[Manandhar \& Sadananda(2002)Manandhar and Sadananda]{manandharEffectHammingDistance2002}
S.K. Manandhar and R.~Sadananda.
\newblock Effect of {{Hamming}} distance of patterns on storage capacity of {{Hopfield}} network.
\newblock In \emph{Proceedings of the 9th {{International Conference}} on {{Neural Information Processing}}, 2002. {{ICONIP}} '02.}, volume~1, pp.\  253--256 vol.1, November 2002.
\newblock \doi{10.1109/ICONIP.2002.1202172}.

\bibitem[Ramsauer et~al.(2020)Ramsauer, Sch{\"a}fl, Lehner, Seidl, Widrich, Adler, Gruber, Holzleitner, Pavlovi{\'c}, Sandve, Greiff, Kreil, Kopp, Klambauer, Brandstetter, and Hochreiter]{ramsauerHopfieldNetworksAll2020}
Hubert Ramsauer, Bernhard Sch{\"a}fl, Johannes Lehner, Philipp Seidl, Michael Widrich, Thomas Adler, Lukas Gruber, Markus Holzleitner, Milena Pavlovi{\'c}, Geir~Kjetil Sandve, Victor Greiff, David Kreil, Michael Kopp, G{\"u}nter Klambauer, Johannes Brandstetter, and Sepp Hochreiter.
\newblock Hopfield {{Networks}} is {{All You Need}}, July 2020.

\bibitem[Shannon(1948)]{shannonMathematicalTheoryCommunication1948}
Claude~Elwood Shannon.
\newblock A mathematical theory of communication.
\newblock \emph{The Bell system technical journal}, 27\penalty0 (3):\penalty0 379--423, 1948.

\bibitem[Vaswani(2017)]{vaswaniAttentionAllYou2017}
A.~Vaswani.
\newblock Attention is all you need.
\newblock \emph{Advances in Neural Information Processing Systems}, 2017.

\bibitem[Widrich et~al.(2020)Widrich, Sch{\"a}fl, Pavlovi{\'c}, Ramsauer, Gruber, Holzleitner, Brandstetter, Sandve, Greiff, Hochreiter, and Klambauer]{widrichModernHopfieldNetworks2020}
Michael Widrich, Bernhard Sch{\"a}fl, Milena Pavlovi{\'c}, Hubert Ramsauer, Lukas Gruber, Markus Holzleitner, Johannes Brandstetter, Geir~Kjetil Sandve, Victor Greiff, Sepp Hochreiter, and G{\"u}nter Klambauer.
\newblock Modern {{Hopfield Networks}} and {{Attention}} for {{Immune Repertoire Classification}}, April 2020.

\bibitem[Wu et~al.(2024)Wu, Hu, Hsiao, and Liu]{wuUniformMemoryRetrieval2024}
Dennis Wu, Jerry Yao-Chieh Hu, Teng-Yun Hsiao, and Han Liu.
\newblock Uniform {{Memory Retrieval}} with {{Larger Capacity}} for {{Modern Hopfield Models}}, November 2024.

\end{thebibliography}
\bibliographystyle{nfam2025_workshop}

\clearpage

\appendix
\section{Appendix}

\subsection{Baseline Analysis.}\label{appendix:baseline}

We empirically confirm the known fact that memory capacity $K_{\max}$ scales exponentially with average separation in the dataset \citep{huProvablyOptimalMemory2024}.

\begin{figure}[ht]
    \vspace{0\baselineskip} 
    \centering
    \begin{subfigure}[b]{0.48\textwidth}
        \centering
        \includegraphics[width=\textwidth]{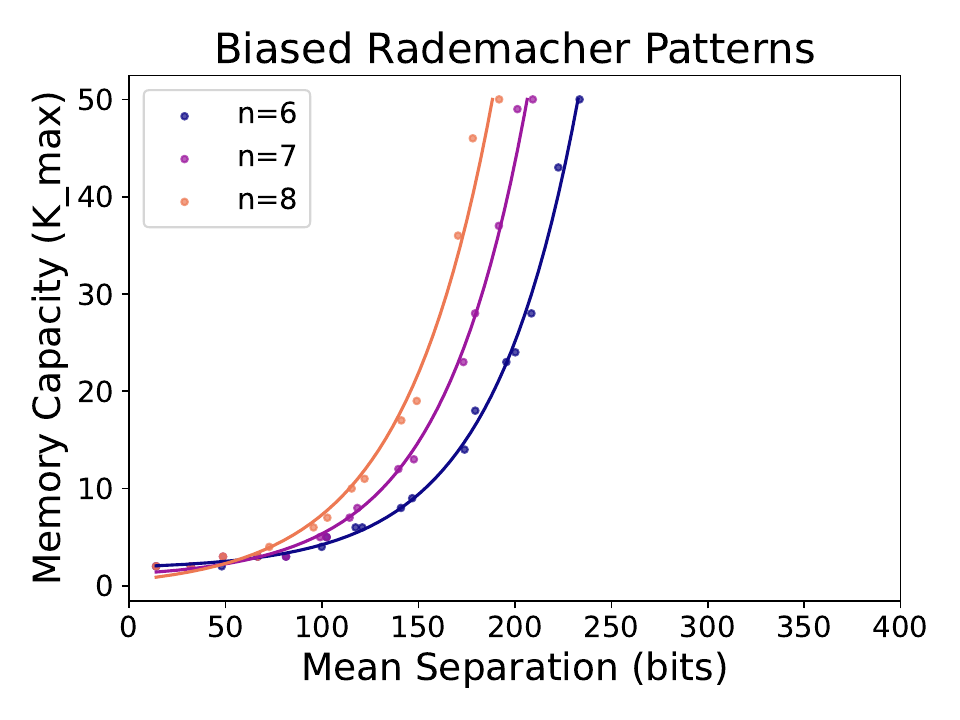}
        \label{fig:artificial}
    \end{subfigure}
    \hfill
    \begin{subfigure}[b]{0.48\textwidth}
        \centering
        \includegraphics[width=\textwidth]{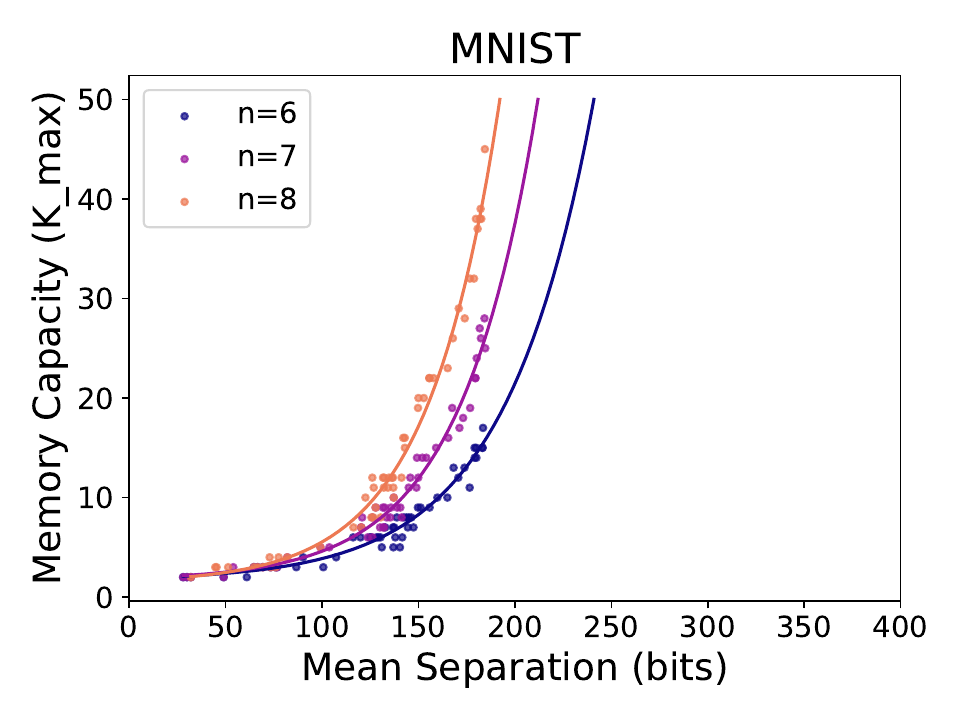}
        \label{fig:mnist}
    \end{subfigure}
    \captionsetup{skip=-8pt} 
    \caption{We calculate memory capacity $K_{max}$ of Dense Associative Memory for 50 sets of biased Rademacher patterns and 53 sets of MNIST of increasing separation. Capacity grows exponentially with data separation for both types. This is in line with \citet{huProvablyOptimalMemory2024}'s theoretical proposition that average separation scales in $O(\ln M)$, where $M$ is $K_{max}$ as memory capacity. When patterns are sufficiently separated, the chance of retrieval error drops dramatically due to lack of crosstalk. The general scaling relationship for feature-correlated data is not different from synthetic data, but there exist important differences which are explained in the results section.}
    \label{fig:ccc}
\end{figure}

\subsection{Experiment and Code.}
All numerical computations were performed using Apple's MLX framework, with patterns and network states represented as 32-bit floating-point arrays.

Our implementation of methods and experiments is publicly available on \href{https://github.com/stefanbielmeier/feature-correlations-am}{GitHub}.

\end{document}